\documentclass[onecolumn]{article}
\usepackage{amsmath}
\usepackage{graphicx}
\usepackage{hyperref}       
\usepackage{url}            
\usepackage[english]{babel}
\usepackage[nottoc]{tocbibind}
\usepackage[square,numbers]{natbib}
\usepackage{authblk}
\usepackage{subfigure}
\usepackage{booktabs}
\usepackage{multirow}
\usepackage{threeparttable}
\usepackage{lscape}

\graphicspath{ {./images/} }

\title{\textbf{MOYU: A Theoretical Study on Massive Over-activation Yielded Uplifts in LLMs}}

\author{Chi Ma}
\author{Mincong Huang}
\author{Chao Wang}
\author{Yujie Wang\thanks{Contact email:wangyujie37@meituan.com}}
\author{Lei Yu}
\affil{Meituan}
\date{}
\begin{document}
\maketitle
\begin{abstract}
Massive Over-activation Yielded Uplifts(MOYU) is an inherent property of large language models, and dynamic activation(DA) based on the MOYU property is a clever yet under-explored strategy designed to accelerate inference in these models. Existing methods that utilize MOYU often face a significant 'Impossible Trinity': struggling to simultaneously maintain model performance, enhance inference speed, and extend applicability across various architectures. Due to the theoretical ambiguities surrounding MOYU, this paper elucidates the root cause of the MOYU property and outlines the mechanisms behind two primary limitations encountered by current DA methods:
1) history-related activation uncertainty, and 
2) semantic-irrelevant activation inertia.
Our analysis not only underscores the limitations of current dynamic activation strategies within large-scale LLaMA models but also proposes opportunities for refining the design of future sparsity schemes.
\end{abstract}

\section{Introduction}\label{section: intro}
Large language models(LLMs), including LLaMA, GPT, and OPT series, have showcased remarkable performance and in-context learning capabilities through the utilization of extensive parameters. However, their computational and memory demands during inference, especially in scenarios sensitive to latency, are substantial. To address these challenges, various techniques based on Massive Over-activation Yielded Uplifts(MOYU) have been proposed. These methods aim to reduce latency in these models by minimizing the excessive activation of heads, neurons, or weights during inference.

Existing MOYU-based techniques can be categorized into \textit{static} and \textit{dynamic activation} methods. Static activation(SA), such as pruning, reduces the excess activated weights in LLMs based on metrics like magnitude, and can be applied either once or iteratively. These configurations remain unchanged for all subsequent inputs and are fully activated during inference. However, a key limitation of SA is that once the process is complete, the deactivated weights cannot be reactivated without undergoing a recovery phase, which could lead to diminished performance and a loss of in-context learning capabilities. Furthermore, the iterative process of SA requires substantial additional training efforts, which may not proportionately enhance the speedup.

\begin{figure}[h]
\centering
\subfigure[TDA\label{figure: tda}]{\includegraphics[height=2in, width=1.9in]{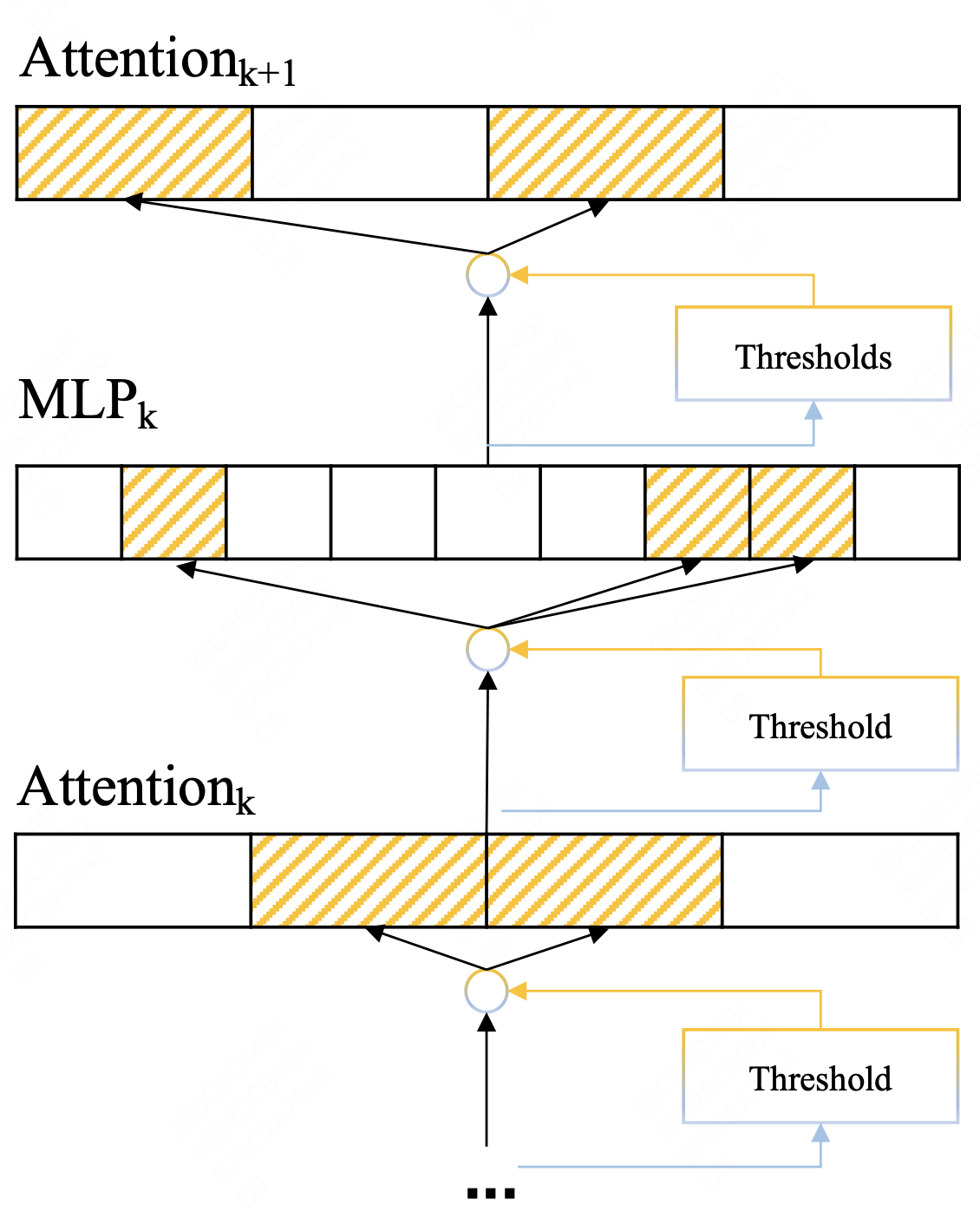}}\hspace{10mm}
\subfigure[DejaVu RODA\label{figure: roda}]{\includegraphics[height=1.9in, width=1.9in]{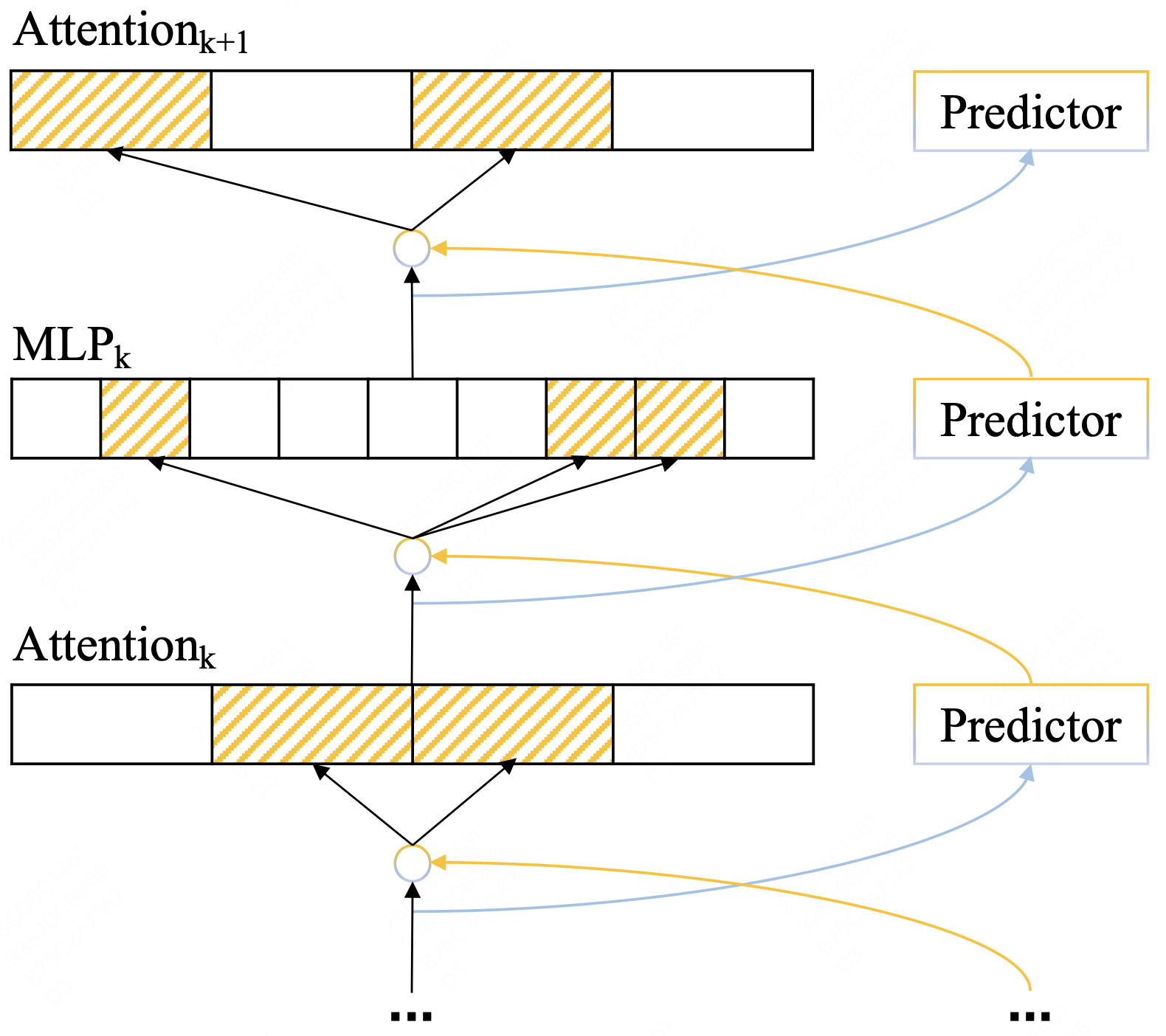}}
\\
\subfigure[MoE RIDA\label{figure: moe rida}]{\includegraphics[height=2in, width=1.9in]{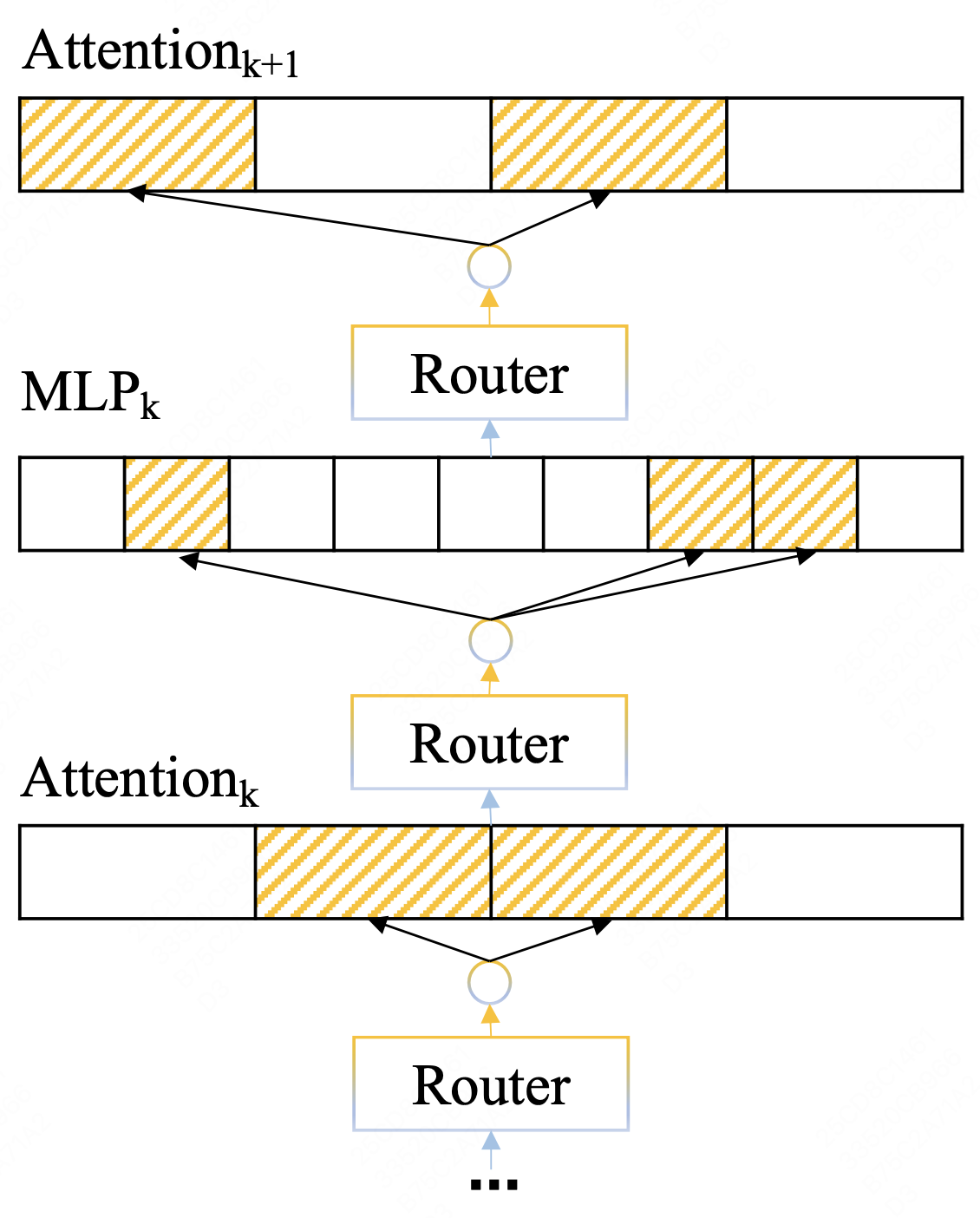}}\hspace{10mm}
\subfigure[Sequential RIDA\label{figure: srida}]{\includegraphics[height=2in, width=1.9in]{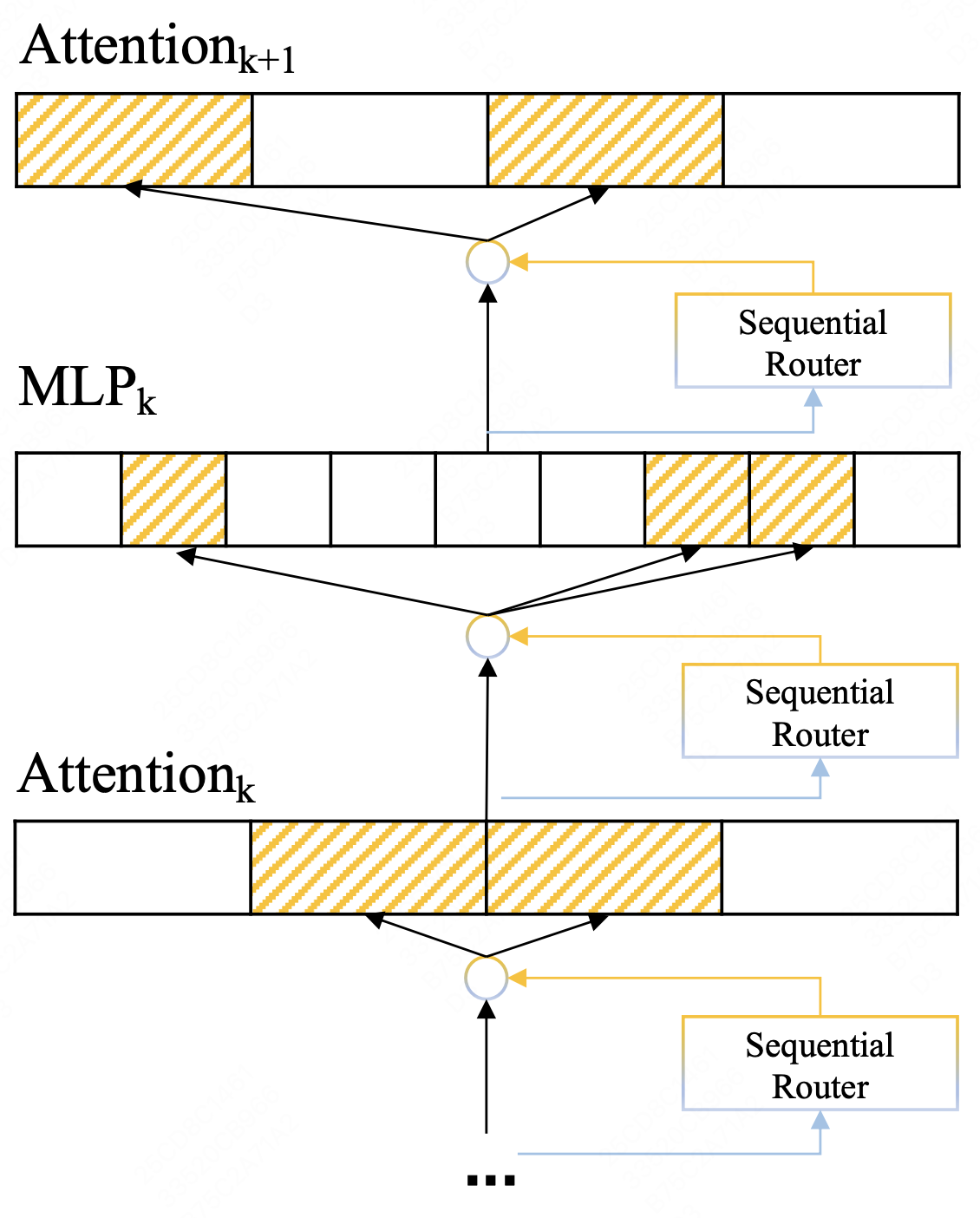}}
\caption{Four kinds of DA methods}
\label{figure: four da}
\end{figure}

On the other hand, MOYU-based dynamic activation(DA) offers adaptability by selectively activating certain heads or neurons during inference, thereby enhancing computational efficiency. This approach capitalizes on the inherent property of massive over-activation found in LLMs to optimize resource utilization. The existing research on DA can be categorized as follows:
\begin{enumerate}
\item
\textbf{Threshold Dynamic Activation(TDA)}: TDA uses a predefined threshold to determine which activation units to retain or discard, as depicted in Figure \ref{figure: tda}. Units with activation values below this threshold are either set to zero or removed during the current forward propagation, thus reducing computational overhead.
\item
\textbf{Router-off-the-loop Dynamic Activation(RODA)}: This method employs a pre-trained \textit{router} block to dynamically determine which activation units are crucial during the model's forward propagation. The router is trained using the model's historical data. For instance, DejaVu\cite{liu2023deja} utilizes a predictive router comprising a two-layer linear network as shown in Figure \ref{figure: roda}
\item
\textbf{Router-in-the-loop Dynamic Activation(RIDA)}: In contrast to RODA, the \textit{router} in RIDA dynamically makes decisions based on the current input and contextual information. This enables the router to adjust its routing strategy in real-time, adeptly managing the complexities of the task at hand, thereby enhancing both efficiency and accuracy. RIDA is primarily implemented within the MoE(Mixture of Experts) structure(in Figure \ref{figure: moe rida}. For example, DS-MoE\cite{pan2024dense} employs a TopK router in its framework. Similarly, Griffin\cite{dong2024promptprompted} also utilizes a TopK router to construct MoE from a dense model in a train-free manner(in Figure \ref{figure: srida}).
\end{enumerate}

Given the 'Impossible Trinity', TDA struggles to enhance inference speed, while RODA fails to extend its applicability between ReLU and non-ReLU activated architectures. MoE-based RIDA necessitates training with the entire model, whereas sequential RIDA, as demonstrated by Griffin, can simultaneously address these three challenges. 

However, despite these advancements, current research on MOYU and DA still lacks a comprehensive theoretical framework that adequately explains the MOYU phenomena across various architectures and activation functions, as well as the underlying mechanisms of MOYU within sequences.

Firstly, we have developed a mathematical rationale that elucidates the origins of the MOYU phenomenon. Then, from this perspective, we have analyzed the causes of two major limitations of existing DA methods:
\begin{itemize}
\item \textbf{Limitation 1:} the restriction to ReLU activation functions.

we suggest that at the \textit{token-level}, history-information-related activation uncertainty(in Section \ref{section: his}) makes it challenging to predict the importance of weights in non-ReLU models, thereby limiting token-level RODA methods to ReLU models.

\item \textbf{Limitation 2:} the inability to identify active neurons based on semantic similarity.

we suggest that at the \textit{sequence-level}, neuron activation is semantically irrelevant(in Section \ref{section: sem}). In other words, neurons are more likely to be activated by the most dominant elements within the same sequence rather than by the semantic content of the input itself, which in turn limits sequence-level DA to RIDA instead of RODA.

\item
In short, it is disheartening that technically, we only have three DA strategies: token-level RODA for ReLU models, token-level RIDA(MoE), and sequence-level TDA and RIDA as discussed in this paper.
\end{itemize}

The rest of the paper is organized as follows. Related works are reviewed in Section \ref{section: related works}. We introduce our universal theoretical framework in Section \ref{section: unveiling} and Section \ref{section: sequencing}, and draw conclusions and limitations in Section \ref{section: conclusion}.

\section{Related Works}\label{section: related works}

\subsection{Massive Over-activation}
In the study of Large Language Models(LLMs), the term \textit{massive over-activation} refers to the excessive activation of numerous neurons during task execution, which can lead to computational waste and decreased efficiency\cite{bommasani2022opportunities, yuan2024llm}. Research\cite{liu2023modelbased} indicates that dense deep neural networks often suffer from this issue. By treating the discrete sparse process as a continuous problem, optimization of the model architecture from end-to-end becomes feasible. The Lottery Hypothesis\cite{frankle2019lottery, malach2020proving} further emphasizes the significance of pruning techniques in reducing unnecessary connections and mitigating over-activation in dense models.

Additional research\cite{shazeer2017outrageously} introduces this concept through a "sparsely-gated mixture-of-experts(MoE) layer," which increases model capacity while reducing computational costs. Moreover, MC-SMoE\cite{li2024merge} tackles the issue of massive over-activation in MoEs by streamlining the model architecture. This is achieved through the merging and low-rank decomposition of redundant experts, guided by the router's information.

\subsection{TDA and RODA}
Research\cite{liu2023modelbased, mirzadeh2023relu} elucidates the capacity of the ReLU to introduce activation sparsity and proposes the concept of dynamic activation. DejaVu\cite{liu2023deja} identifies that the sparsity introduced by ReLU can be predicted, thus proposing the first viable RODA scheme. On the OPT series, DejaVu facilitates a 2-6x acceleration in inference latency at 75\% sparsity. Building upon the DejaVu approach, ReLU$^2$\cite{zhang2024relu2} first applies TDA to non-ReLU models and achieves nearly 70\% sparsity with minimal loss to model performance. ProSparse\cite{song2024prosparse} proposes a practical DA inference framework and, building on ReLU$^2$, achieves only a 1-percent increase in perplexity at approximately 80\% sparsity by replacing the activation function and continuing to induce sparsity.

\subsection{RIDA}
Router-in-the-loop is the predominant method within the MoE framework. Unlike TDA and RODA methods, most RIDA approaches rely on training an expert router to facilitate dynamic activation. 
LLaMA-MoE\cite{llama-moe-2023} transforms feed-forward networks(FFNs) into MoEs by constructing experts and training an additional gating network for expert routing. DS-MoE\cite{pan2024dense} introduces a framework that utilizes dense computation during training and switches to sparse computation during inference, significantly enhancing parameter efficiency over traditional sparse MoE methods and reducing the total parameter count.
Learn-To-be-Efficient\cite{zheng2024learn} achieves an optimal balance between sparsity and performance by activating fewer neurons and is applicable to models with both ReLU and non-ReLU activation functions. Lory\cite{zhong2024lory} retains the autoregressive properties of language models by adopting a causally segmented routing strategy and a similarity-based data batching method. This approach enables efficient expert merging operations and promotes specialization among experts in processing similar documents during training sessions.

\section{Unveiling MOYU}\label{section: unveiling}
Section \ref{section: related works} provides a review of the literature relevant to MOYU. This section begins by outlining the theoretical foundations of MOYU and then presents mathematical proof. Following literature\cite{li2023lazy}, we can demonstrate through the following derivation how massive over-activation arises and why SwiGLU cannot produce greater sparsity than ReLU. 

Assuming a neural network as in Equation~\ref{eq: initiate}:
\begin{equation}
    f(x)=\boldsymbol{V}\sigma (p(\boldsymbol{x};\boldsymbol{\theta}))
\label{eq: initiate}
\end{equation}
,where \(\boldsymbol{V} = [v_1, ..., v_{d_{ff}}]\) is network parameter for the last layer drawn from a random distribution, \(\sigma()\) is the SwiGLU activation function, and \(p(\boldsymbol{x};\boldsymbol{\theta})\) denotes all other layers with parameter \(\theta\). We write \(p = p(\boldsymbol{x};\boldsymbol{\theta})\) for simplicity. 

Consider the cross-entropy(CE) loss with function \(\ell_{CE}(f(\boldsymbol{x}),\boldsymbol{y})\), where \(\boldsymbol{y}\) is an arbitrary vector that sums up to one and independent of \(\boldsymbol{V}\). Assume that the entries of \(\boldsymbol{V}\) are drawn from independent distributions, the probability of any entry of \(\boldsymbol{V}\) being 0 is less than 1, and \(E[\boldsymbol{V}] = 0\) . If there exist an \(i^*\) such that \(p_{i^*} > 0\), then we have Equation \ref{eq: lianshiqiudao}:
\begin{equation}
    \frac{\partial \ell}{\partial p_{i*}} =\left \langle \frac{\partial \ell}{\partial f}, \frac{\partial f}{\partial p_{i*}} \right \rangle=\left \langle \frac{\partial \ell}{\partial f},v_{i^*} \right \rangle  
\label{eq: lianshiqiudao}
\end{equation}

Substituting CE loss function into Equation \ref{eq: lianshiqiudao} yields Equation \ref{eq: lianshiqiudao first term}:
\begin{equation}
\begin{aligned}
    \frac{\partial \ell_{CE}}{\partial f}
    &=\frac{exp(f(x))}{\left \langle exp(f(x)),\boldsymbol{1} \right \rangle }-y\\
    &=\frac{exp( {\textstyle \sum_{i}\sigma(p_i)\cdot \boldsymbol{v_i}})}{\left \langle exp( {\textstyle \sum_{i}\sigma(p_i)\cdot \boldsymbol{v_i}}), \boldsymbol{1}\right \rangle}-y
\end{aligned}
\label{eq: lianshiqiudao first term}
\end{equation}

By substituting Equation \ref{eq: lianshiqiudao first term} back into Equation \ref{eq: lianshiqiudao}, we can obtain Equation~\ref{eq: lianshiqiudao second term}:
\begin{equation}
\begin{aligned}
    \frac{\partial \ell_{CE}}{\partial p_{i^*}}
    &=\frac{\left \langle exp(\sum_{i}\sigma(p_i) \cdot \boldsymbol{v_i} ),\boldsymbol{v_{i^*}} \right \rangle }{\left \langle exp(\sum_{i}\sigma(p_i) \cdot \boldsymbol{v_i} ),\boldsymbol{1} \right \rangle }- \left \langle \boldsymbol{v_{i^*}},y \right \rangle 
\end{aligned}
\label{eq: lianshiqiudao second term}
\end{equation} 

Expanding the numerator of Equation \ref{eq: lianshiqiudao second term} yields Equation \ref{eq: enumerator}. In Equation\ref{eq: enumerator}, we assume that parameter \( \theta \) and \( \tau \) have no negative features. If we have \(p_{i^*}^0=Swish_1(x\theta)\odot(x\tau)\) and \(p_{i^*}^1=ReLU(x)\) respectively, it is easy to get \(Swish_1(x\theta) < x\theta\) when \(x>0\), and \(p_{i^*}^0 < x\theta = p_{i^*}^1\) and \(p_{i^*}^0 < x\tau\) holds true.
\begin{equation}\footnotesize
\begin{aligned}
    \left \langle exp(\sum_{i}\sigma(p_i) \cdot \boldsymbol{v_i} ),\boldsymbol{v_{i^*}} \right \rangle 
    &= \sum_{m}(v_{i^*,m} \cdot exp(\sum_{i}\sigma(p_i) \cdot v_{im}) \\
    &= \sum_{m}(v_{i^*,m} \cdot exp(p_{i^*} \cdot v_{i^*m}) \cdot exp(\sum_{i \neq i^*}\sigma(p_i) \cdot v_{im}) 
\end{aligned}
\label{eq: enumerator}
\end{equation} 

Similar to literature\cite{li2023lazy}, we also have \(\mathrm {E}[\frac{\partial \ell_{CE}}{\partial p_{i^*}}] > 0\) holds true since the expectation of \textbf{V} is zero and the transformation of the activation function does not change the non-negative property of the loss expectations. 
\begin{equation}
\begin{aligned}
    \mathrm {E}[\frac{C_1V \cdot exp(pV)}{C_2~exp(pV)+C_3}] =\mathrm {E}[\frac{C_1V}{C_2 + C_3exp(-pV)} ]
\end{aligned}
\label{eq: loss expectation}
\end{equation} 

The first term on the right-hand side (RHS) of the loss function (as shown in Equation \ref{eq: lianshiqiudao second term}) can be simplified to the form presented in Equation \ref{eq: loss expectation}, while the expectation of the second term on the RHS is zero. Given \(p_{i^*}^0 < p_{i^*}^1\), Equation~\ref{eq: loss expectation} demonstrates that switching the activation function from ReLU to SwiGLU \textbf{decreases} the expected value of the loss function.

This implies that if there exists an \(i^*\) such that \(p_{i^*} > 0\), the gradient of the cross-entropy loss with respect to any positive activation \(p_{i^*} > 0\) is positive in expectation. Consequently, any training algorithm that follows the negative gradient direction tends to \textbf{reduce} the magnitude of such positive activation, leading to a smaller training loss and thus promoting \textbf{sparsity}.

In this process, \textbf{the ReLU activation function causes a greater reduction in magnitude compared to SwiGLU}.

\section{Sequencing MOYU}\label{section: sequencing}
In Section \ref{section: unveiling}, this paper theoretically deduces the root causes of the MOYU phenomenon and explores how non-ReLU activation functions might mitigate it. The literature\cite{georgiadis2019accelerating, kurtz2020pmlr, zhu2023cvprw} has highlighted that the current level of activation map sparsity is insufficient to fully exploit the performance of DA methods. In this section, we identify two limitations associated with choosing DA methods as discussed in Sections \ref{section: his} and \ref{section: sem}.

\subsection{History-related Activation Uncertainty}\label{section: his}
RODA schemes excel in models that utilize ReLU as the activation function\cite{mirzadeh2023relu,liu2023deja, zhang2024relu2, song2024prosparse}. However, in models employing non-ReLU activation functions, the offline-trained router struggles to accurately select which heads and neurons will be activated\cite{ma2024dynamic, dong2024promptprompted}.

We suggest in this section that the failure of RODA in non-ReLU scenarios is closely linked to shifts in weight importance under different historical inputs: a router trained on diverse historical activation data may find it challenging to accurately identify the weights that are most crucial for the current

Similarly, we assume the presence of a ReLU-activated model as described in Equation \ref{eq: initiate}. And the simplified loss of input token \(x_i\) can be described as(Equation \ref{eq: model loss}):
\begin{equation}\footnotesize
\begin{aligned}
    L_i=(\frac{\partial f}{\partial x_i}\mathrm{d}x_i + \frac{\partial f}{\partial \mathbf{\theta}_i } \mathrm{d}\mathbf{\theta}_i)^T(\frac{\partial f}{\partial x_i}\mathrm{d}x_i + \frac{\partial f}{\partial \mathbf{\theta}_i } \mathrm{d}\mathbf{\theta}_i)
\end{aligned}
\label{eq: model loss}
\end{equation}

Weight change sensitivity(gradients) in model training is as Equation \ref{eq: output qiudao}:
\begin{equation}
\begin{aligned}
    \frac{\partial L_i}{\partial \mathrm{d}\mathbf{\theta}_i} = 2(\frac{\partial f}{\partial x_i}\mathrm{d}x_i + \frac{\partial f}{\partial \mathbf{\theta}_i } \mathrm{d}\mathbf{\theta}_i)\frac{\partial f}{\partial \mathbf{\theta}_i}
\end{aligned}
\label{eq: output qiudao}
\end{equation}

By summing gradients, we have Equation \ref{eq: gradient sum}:
\begin{equation}
\begin{aligned}
    \nabla _{\mathrm{d}\theta_i}L 
    &= \sum_{i}2(\frac{\partial f}{\partial x_i}\mathrm{d}x_i + \frac{\partial f}{\partial \mathbf{\theta}_i } \mathrm{d}\mathbf{\theta}_i)\frac{\partial f}{\partial \mathbf{\theta}_i}\\
    &= \nabla _{\mathrm{d}\theta_i}L_i+\sum_{j=0:i-1}\nabla _{\mathrm{d}\theta_j}L_{j}
\end{aligned}
\label{eq: gradient sum}
\end{equation}

And the importance of model weights can be described in Equation \ref{eq: weight importance}:
\begin{equation}
\begin{aligned}
    \Theta_i 
    &= \sum_{i}|V \cdot \nabla _{\mathrm{d}\theta_i}L_i| \\
    &=|V|\cdot \sum_{i}|\nabla _{\mathrm{d}\theta_i}L_i|\\
    &=|V |\cdot (\nabla _{\mathrm{d}\theta_i}L_i+\sum_{j=0:i-1}\nabla _{\mathrm{d}\theta_j}L_{j})\\
    &= |V |\cdot \nabla _{\mathrm{d}\theta_i}L_i+\Theta_{i-1}
\end{aligned}
\label{eq: weight importance}
\end{equation}
, which means weight importance of a model are not only related to current input along the direction of \(\theta\), but also to the cumulative gradient information from all previous data. 

For models utilizing ReLU activation, Equation \ref{eq: weight importance} simplifies to the sum of the weights corresponding to positive inputs, which linearly correlates with the magnitude of the current weights themselves. However, for models employing non-ReLU activations, the significance of the current weights becomes considerably more complex.

\subsection{Semantic-irrelevant Activation Inertia}\label{section: sem}
Using a simplified loss function, Section \ref{section: his} demonstrated that models with non-ReLU activation rely on historical information to accurately decide which neurons to activate. This section reveals that historical information is significantly influenced by the Heavy Hitter (\(H_2\)), and the occurrence of \(H_2\) is not related to semantics\cite{sun2024massive}.

Following literature\cite{zhang2023h2o} we have \(H_2: S^*\subset [m]\), and \(k = |S^*|,~\tau \in (0, 1)\) denote a threshold. \(\alpha \in (0,1)\) denote a fraction of mass(larger than \(\tau\)) outside \(S^*\).

It is natural that attention with \(H_2\) is a \((\alpha, \tau, k)\)-good mapping since for all \(x \in \textbf{R}^d\), \(S^* \subset supp_\tau(Att(x))\), and \(|supp_\tau(Att(x))\setminus S^*| \le \alpha \cdot k\). Then we have \(S^* \subseteq \cap_{i\in[n]}supp_\tau(x_i)\), and \(|(\cup_{i\in[n]}supp_\tau(Att(x)))\setminus S^*| \le \alpha k n\) for \(x_i\) draw from \((\alpha, \tau, k)\)-good distribution uniformly at random. \textbf{That is to say, \(H_2\) in a sequence significantly decides the activation pattern.}

\begin{figure}[]
\centering
\subfigure[Neuron activation pattern of tokens sampled from a single sentence \& input in parallel\label{figure: sent par}]{\includegraphics[width=2.7in]{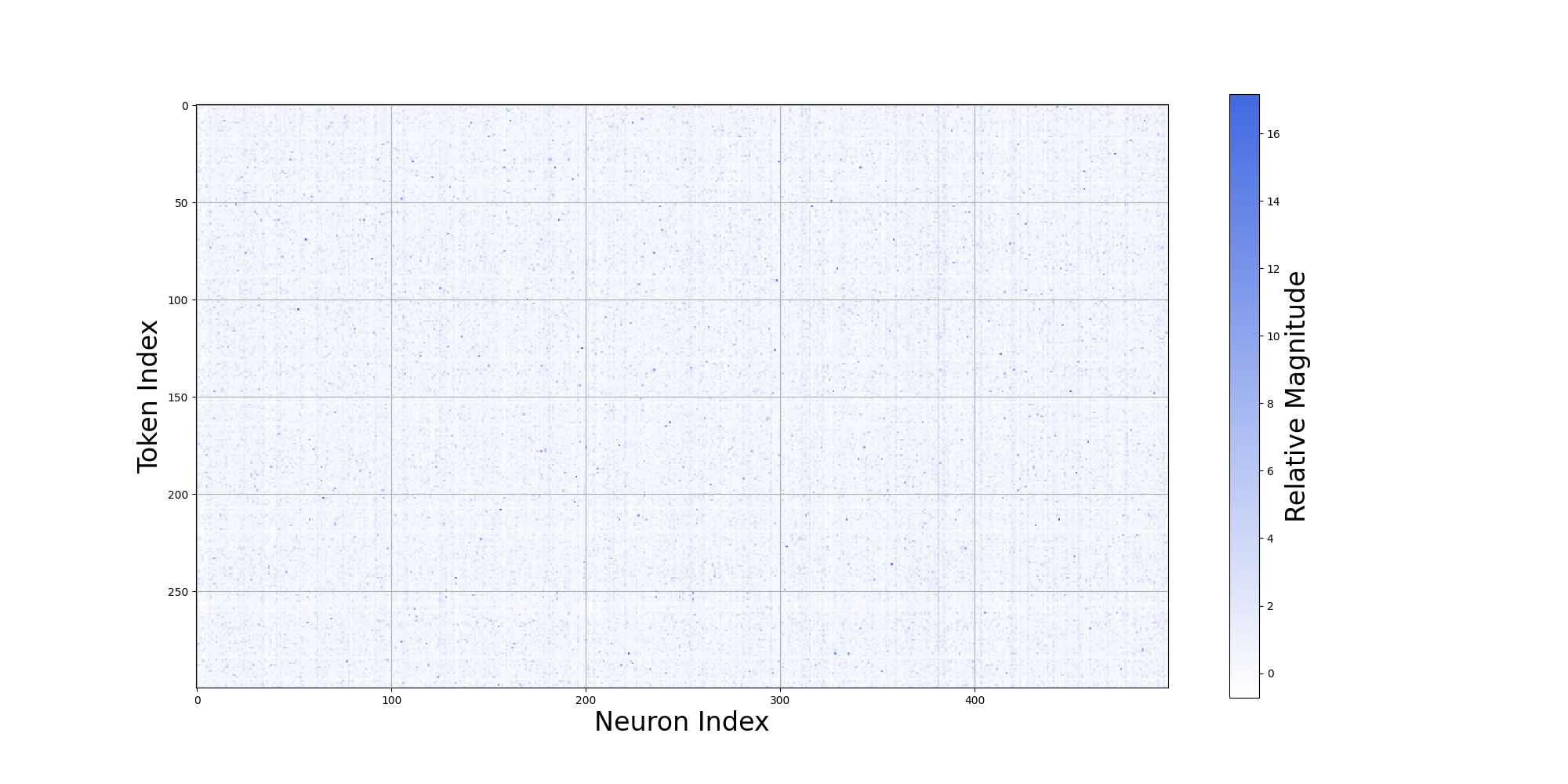}}
\subfigure[Neuron activation pattern of tokens sampled from a single sentence \& input sequentially\label{figure: sent seq}]{\includegraphics[width=2.7in]{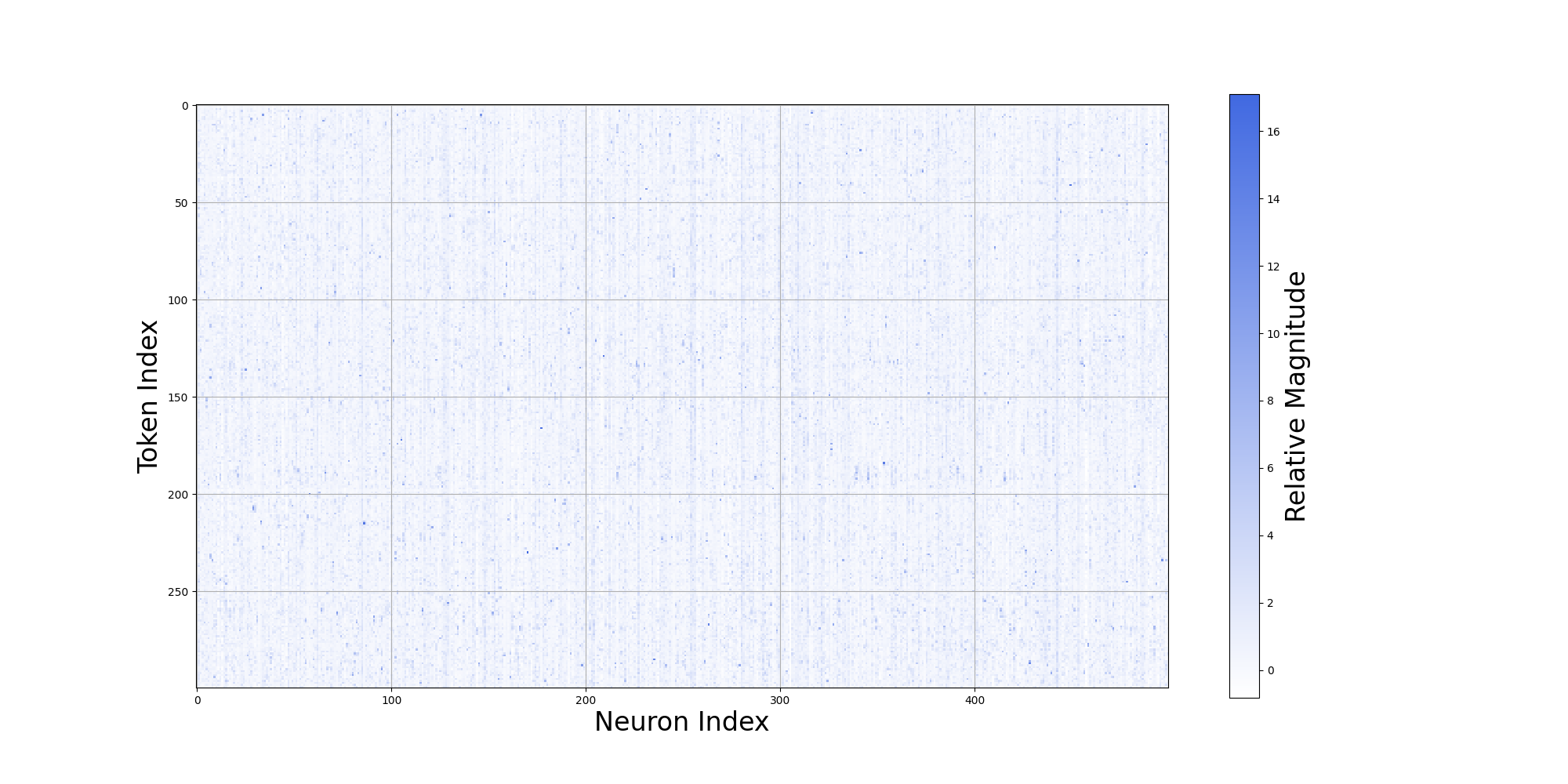}}
\\
\subfigure[Neuron activation pattern of random tokens \& input in parallel\label{figure: rand par}]{\includegraphics[width=2.7in]{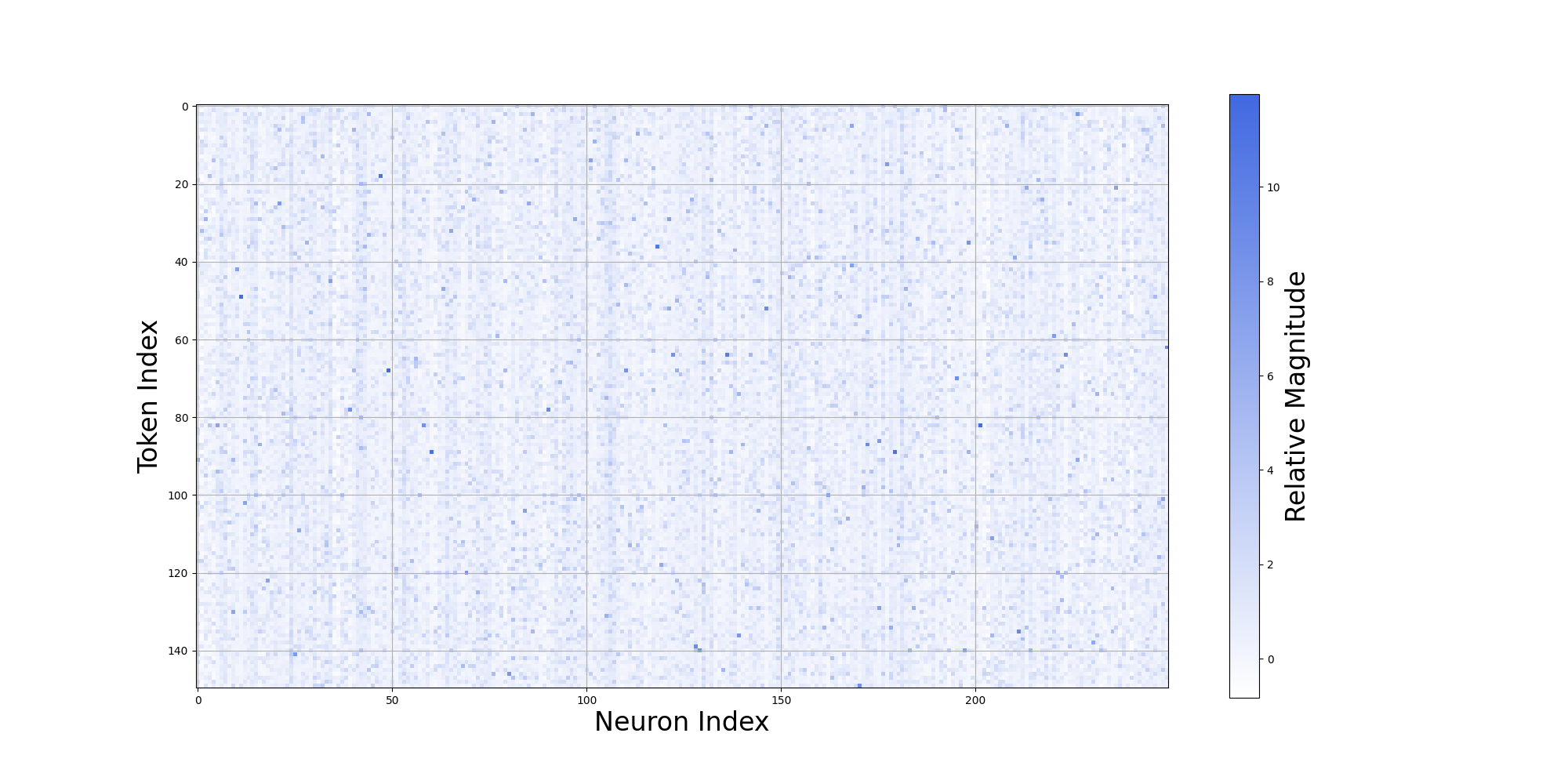}}
\subfigure[Neuron activation pattern of random tokens \& input sequentially\label{figure: rand seq}]{\includegraphics[width=2.7in]{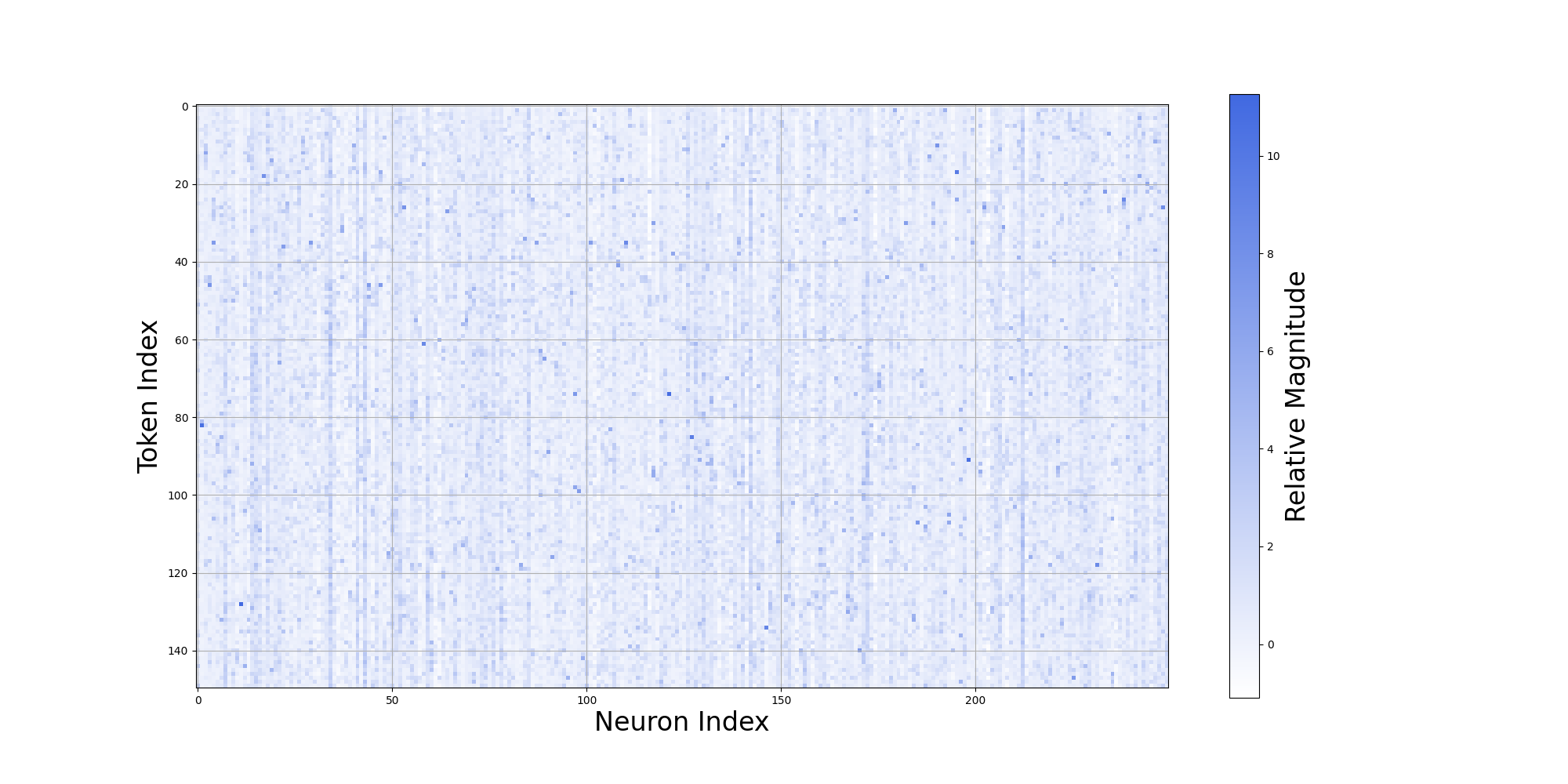}}
\caption{Neuron Activation Pattern Comparisons Across Different Sampling and Input Manners}
\label{figure: four activation patterns}
\end{figure}

Figures \ref{figure: sent par} through \ref{figure: rand seq} showcase the phenomenon of activation inertia and its lack of semantic relevance. Figures \ref{figure: sent par} and \ref{figure: sent seq} depict the activation pattern of neurons when tokens from a single sentence are input either individually or sequentially. Conversely, Figures \ref{figure: rand par} and \ref{figure: rand seq} reveal the neuron activation when tokens from a random word list are introduced in the same formats.

The horizontal axis in Figure \ref{figure: sent par} to Figure \ref{figure: rand seq} represents the neuron index, while the vertical axis represents the token index. The colors in the figures indicate the relative activation values for each token by each neuron. To make the data between different neurons and tokens comparable and to clarify the images, the activation values have been normalized in this study. From these images, we can get the following insights:
\begin{enumerate}
    \item Sequential input induces activation inertia.
    
    Compared to Figure \ref{figure: sent par}, the narrow vertical lines in Figure \ref{figure: sent seq} are more pronounced. A similar pattern is observed when comparing Figure \ref{figure: rand par} with Figure \ref{figure: rand seq}. This suggests that for tokens from the same source, when inputs are made sequence, or as referred to by Griffin\cite{dong2024promptprompted} as a "prompt" into the model, the neurons activated by these input tokens will exhibit a clear clustering phenomenon. Neurons activated by the previous token demonstrate greater activation inertia and are thus more likely to be activated by subsequent tokens.
    
    \item Activation inertia is semantic-irrelevant.
    
    The narrow vertical lines in Figure \ref{figure: rand seq} are more pronounced compared to those in Figure \ref{figure: sent seq}. This comparison indicates that the activation inertia phenomenon observed is more intense with tokens from a random word list, confirming that activation inertia is semantically irrelevant, as highlighted in the title of this section.
    
    The intuitive reason behind this, as the paper posits, follows the theoretical analysis at the beginning of this section: \textbf{activation patterns are caused by the earliest heavy hitters and maintained by semantically irrelevant activation inertia}. The proportion of words with actual semantics in a random word list is higher than in an English sentence with many prepositions and articles, thus more significantly causing and maintaining activation patterns.
    
    \item
    
    There is no significant difference in the narrow vertical line patterns between Figure \ref{figure: sent par} and Figure \ref{figure: rand par}. Unlike the conclusions drawn from the previous two sets of observations, this comparison does not involve inputting tokens as a sequence. We suggest that this phenomena aligns with the theoretical derivation at the beginning of this subsection. When activation inertia is at the token level, each token acts as its own heavy hitter, leading to a more diversified activation pattern.
\end{enumerate}

Through Figures \ref{figure: sent par} to \ref{figure: rand seq}, we have confirmed claims that sequential input strengthens activation inertia. It might be evident that activation inertia occurs with sequential input rather than parallel input, given that attention mapping inherently processes all words in sequence.

However, it is crucial to recognize that this activation pattern, initiated by the earliest heavy hitter and sustained by activation inertia, can persist into subsequent generative processes. By combining sequential input with the RIDA method, it is feasible to precisely identify neurons that require activation during subsequent generation phases, with minimal impact on model performance, thereby realizing a training-free RIDA approach.

Griffin's experiments have already confirmed the correctness of this speculation. Future ablation experiments could further test the correctness of the theory in this subsection by examining the impact on model performance of removing the first heavy hitter in a sentence (rather than all, which would be equivalent to excluding all neurons that need to be activated).

\section{Conclusion and Limitations}\label{section: conclusion}
Massive Over-activation Yielded Uplifts(MOYU) are intrinsic characteristics of large language models, and leveraging these properties through Dynamic Activation(DA) is a promising yet underutilized strategy to enhance inference speeds in these models. Traditional methods that exploit MOYU often encounter significant challenges, including maintaining model performance, speeding up inference, or extending their use to various architectures. This paper have developed a mathematical framework that elucidates the origins of the MOYU phenomenon. Through this framework, we have identified two primary limitations of current DA methods: 1) their reliance on ReLU activation functions; 2) their inability to detect active neurons based on semantic similarities.

This paper has following limitations: firstly, the mathematical rationale and implementation of the proposed DA methods could introduce complexities that might impede their practical application. Additionally, this paper highlights that sequence-level activation is predominantly influenced by heavy hitters within the same sequence; however, due to effort constraints, ablation experiment is not conducted. It is anticipated that future research will undertake more extensive experiments.

\bibliographystyle{unsrt}
\bibliography{ref}

\clearpage

\end{document}